\documentclass{ieeeaccess}
\usepackage{cite}
\usepackage{amsmath,amssymb,amsfonts}
\usepackage{algorithmic}
\usepackage{graphicx}
\usepackage{textcomp}
\def\BibTeX{{\rm B\kern-.05em{\sc i\kern-.025em b}\kern-.08em
    T\kern-.1667em\lower.7ex\hbox{E}\kern-.125emX}}
\begin{document}
\history{Date of publication xxxx 00, 0000, date of current version xxxx 00, 0000.}
\doi{10.1109/ACCESS.2017.DOI}

\title{Improving Performance of Object Detection using the Mechanisms of Visual Recognition in Humans}
\author{\uppercase{Amir Ghasemi},
\uppercase{Nasrin Bayat},
\uppercase{Fatemeh Mottaghian},
\uppercase{Akram Bayat}\authorrefmark{1}}

\address[1]{The Berkeley Institute for Data Science, Berkeley, California, USA}

\markboth
{Author \headeretal: Preparation of Papers for IEEE TRANSACTIONS and JOURNALS}
{Author \headeretal: Preparation of Papers for IEEE TRANSACTIONS and JOURNALS}


\begin{abstract}
Object recognition systems are usually trained and evaluated on high resolution images. However, in real world applications, it is common that the images have low resolutions or have small sizes. In this study, we first track  the performance of the state-of-the-art deep object recognition network, Faster-RCNN, as a function of image resolution. The results reveals negative effects of low resolution images on recognition performance. They also show that different spatial frequencies convey different information about the objects in recognition process. It means multi-resolution recognition system can  provides better insight into optimal selection of features that results in better recognition of objects. This is similar to the mechanisms of the human visual systems that are able to implement multi-scale representation of a visual scene simultaneously. Then, we propose a multi-resolution object recognition framework rather than a single-resolution network. The proposed framework is evaluated on the PASCAL VOC2007 database. The experimental results show the performance of our adapted multi-resolution Faster-RCNN framework outperforms the single-resolution Faster-RCNN on input images with various resolutions with an increase in the mean Average Precision (mAP) of 9.14\% across all resolutions and 1.2\% on the full-spectrum images. Furthermore, the proposed model yields robustness of the performance over a wide range of spatial frequencies.
\end{abstract}


\titlepgskip=-15pt
\maketitle
\section{Introduction}
Recent advances in deep neural networks (DNN) \cite{10012263,bayat2017deriving,bayat2021automated,bayat2018scene} and access to very large datasets with million annotated data especially for computer vision applications have led to state-of-the-art results in many problem domains such as object detection and scene classification \cite{zhou2014learning, girshick2014rich, bayat2022human}. For example, Faster-RCNN network \cite{ren2016faster} achieved the impressive results in recognition and localization of objects in natural scenes or GoogleNet \cite{szegedy2015going} reached approximately to the human performance in classification of the ImageNet database \cite{deng2009imagenet}.

Despite the significant achievements, a major drawback of the current deep neural networks for visual recognition is that they have been trained and evaluated on high quality images in which their performances drop significantly in classification of the low resolution images \cite{dodge2016understanding}, \cite{kim2022hair}. 
 
 in other words, previous works only focus on full spectrum image resolutions when training their networks, but the variety of real-world applications, from moving objects to small size images, often demand different constraints.
 Given the real-world resource constraints such as low resolution and small size images, recognition efficiency and performance becomes increasingly important for object detection. However to achieve better efficiency, accuracy usually is scarifies.
This paper aims to tackle this problem by systematically studying the performance of the object recognition networks (e.g., Faster-RCNN) under the various resolutions of input images. 
Then, we propose a method based on the mechanism of the human visual system that produces better and robust performance in deep object detection networks with both higher accuracy and better efficiency across a wide rang of image resolutions. 
To evaluate the performance of our proposed method for object recognition, we choose the Faster-RCNN network \cite{ren2016faster} as one of the best existing models in terms of the accuracy on PASCAL VOC \cite{everingham2010pascal} image database.

\section{Related Work}
\textbf{Object Detection in Human Visual System -} Visual perception in humans is a process that human acquire knowledge about their environment. This process is initiated when surrounding light enters the eye and induces electrical signals subsequently processed within the brain where an image is formed. 

A large volume of studies has shown that human visual system has the ability to adapt to changes in environment in different ways, in which each adjustment may need different mechanisms \cite{webster2015visual,mcgovern2012perceptual, harris2012generalized, webster2011adaptation}. For example adaptation to color encompasses different adjustments including sensitivity changes in the cones.

Human visual system has also spatial adaptations. As an example, tilt aftereffects can be deduced with both real and subjective contours, with asymmetries between them which encourages adaptation at different cortical sites \cite{howard2011mccollough}.

Visual objects in the real world are observed in contextual scenes which are usually relevant from physical and semantic perspective. With regard to blurriness, Bar \textit{et al.} \cite{bar2004visual} deployed a blurred, low frequency representation of a scene and showed that human visual system is able to determine ambiguous objects.\\
\textbf{Object Detection in Computer Vision -} It deals with discovering instances of semantic objects of a certain class (e.g. buildings, cars, or humans) in images and videos. Most traditional approaches for object detection used well-established computer vision methods which relies on extracting feature descriptors (e.g., SIFT, SURF, BRIEF, etc.)\cite{lowe1999object, bay2006surf,calonder2010brief}. However, with the emerge of deep neural networks (in particular, convolutional neural network \cite{szegedy2015going}) and its remarkable success in computer vision, the majority of recent works in object detection for digital images and videos have shifted towards using them as the primary technique \cite{westlake2016detecting}. The state-of-the-art technique using DNN can be categorized into two main types: one-stage methods and two stage-methods. One-stage methods prioritize inference speed (e.g., YOLO \cite{redmon2016you}, SSD \cite{liu2016ssd} and RetinaNet \cite{lin2017focal}). Two-stage methods prioritize detection accuracy (e.g., Faster R-CNN \cite{ren2016faster}, Mask R-CNN \cite{he2017mask} and Cascade R-CNN \cite{cai2018cascade}).

In this study, we focus on improving detection accuracy and efficiency in detecting objects in low resolution images and videos. As said before, the existing neural networks are generally trained and tested on high quality images, in which when they are fed in with low quality images, their performance in detecting and recognizing objects reduces remarkably \cite{kannojia2018effects}. However, in real life, there are many cases that images have low resolution. To solve this issue, motivated by the capability of human visual system in adapting to different range of resolution, we propose a multi-resolution method which improves the performance of the model which it is fed in with blurry images.

\section{Tracking Object recognition network performance}
In this section, we explain our methodology to track performance of the state of the art deep object recognition network, faster-RCNN with various resolution levels along with notation and problem formulation.
\subsection{Model Architecture}

We deploy Faster-RCNN framework for object detection and localization. The Faster-RCNN uses Region Proposal Network (RPN) and an object detection network that share convolutional layers for fast testing. The baseline network is VGG-16 and its Conv 5-3 features are used for region proposal. We adapted the Faster-RCNN based on the publicly available code in \cite{chen2017implementation} 
and implemented it with a few modifications in TensorFlow to evaluate the performance of the network on 4952 images of the PASCAL VOC2007 test dataset with full resolution (full spectrum in the frequency domain).
\subsection{Notation and Problem Formulation}
As said before, we evaluate the performance of the Faster-RCNN network on the PASCAL VOC2007 image database under multiple resolutions. For this purpose, we create a set of image databases whose resolutions vary systematically from extremely coarse to very fine. This is performed by applying the two-dimensional Gaussian low pass filter in various cut off frequencies to the PASCAL VOC2007 image database in the frequency domain. The results of this process resemble blurring of an image to reduce the details of high frequency components of that image in multiple levels. To simplify, instead of applying a two dimensional (2D) Gaussian function to each pixel of an image which is equivalent to convolving a 2D Gaussian with the image, we apply the product of their individual Fourier transforms. We then re-transform the resulting product into the spatial domain to obtain the image in the desired resolution ($I_{f_c}$).

Attenuating frequencies using low pass Gaussian filters \cite{bayat2022white} results in a smoother image in the spatial domain. This process is formalized in the following equations:
\begin{equation}
\hat{F}_{f_c}(u,v)=F(u,v)H_{f_c}(u, v),  f_{min}<f_c<f_{max}
\end{equation}
\begin{equation}
H_{f_c}(u, v)=  e^{-\frac{u^2 + v^2}{2f_c^2}}
\end{equation}
\begin{equation}
I_{f_c} =  f^{-1}\left[\hat{\mathcal{F}}_{f_c}(u, v)\right]
\end{equation}

where $F(u,v)$ is the Fourier transform of the full-spectrum image $I$ and $u,v$ are representative of a particular spatial frequency contained in the spatial domain of image $I$. For instance, $F(0,0)$ represents the DC-component of the image which corresponds to the average brightness. $H_{f_c}$ is the 2D Gaussian spatial filter with cut-off frequency of $f_c$ ranging from $f_{min}$ to $f_{max}$ which is defined systematically between the center and the edge of the Fourier image, $F$ , as follows:
\begin{equation}
f_c =  c .\frac{w}{20}
\end{equation}
In this way, for a given image $I_{w \times h}$:
\begin{equation}
f_{min} =  \frac{S}{20}, f_{max}=S, S=max(w, h)
\end{equation}
By doing so, we systematically create 20 image databases from the PASCAL VOC2007 database in twenty-scale resolutions, namely PASCAL VOC2007- $R_1$, . . . , PASCAL VOC2007- $R_{20}$.
\begin{figure}[t]
\begin{center}
\includegraphics[width=0.48\textwidth]{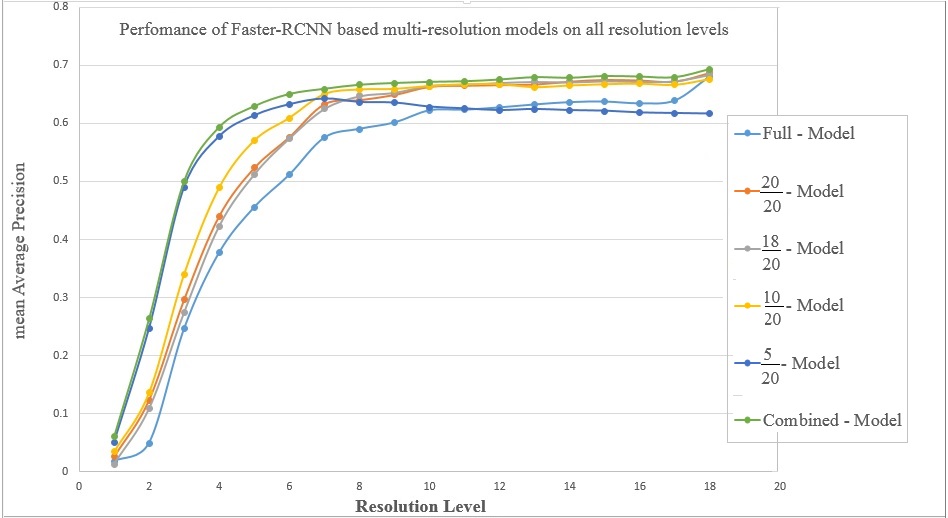}
\end{center}
\caption{The performance of multi-resolution Faster- RCNN obtained from the combination of 5 models on 5 resolution levels ($R_5, R_{10}, R_{18}, R_{20}$, and Full-spectrum). All models have been evaluated on multi-resolution test databases ($R_1, ..., R_{20}$)
}
\label{fig: The performance of multi-resolution}
\end{figure}
\section{Performance V.S. Resolutions}
However, the state of the art object detectors are trained and tested on high resolution images but a most important question is how we can have detectors that give us the best balance of resolution and accuracy for different application needed. In this section, we evaluate the comparison of accuracy v.s. resolution tradeoff.

\begin{figure*} 
    \begin{minipage}[t]{0.325\linewidth}
    \includegraphics[width=\linewidth]{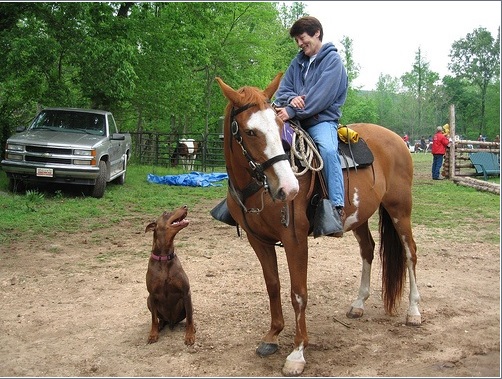}
    \end{minipage}\hfill
    \begin{minipage}[t]{0.325\linewidth}
    \includegraphics[width=\linewidth]{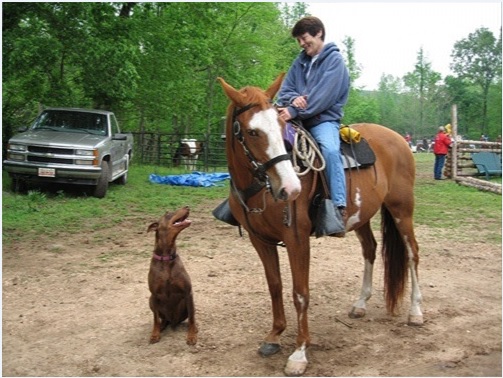}
    \end{minipage}\hfill
    \begin{minipage}[t]{0.325\linewidth}
    \includegraphics[width=\linewidth]{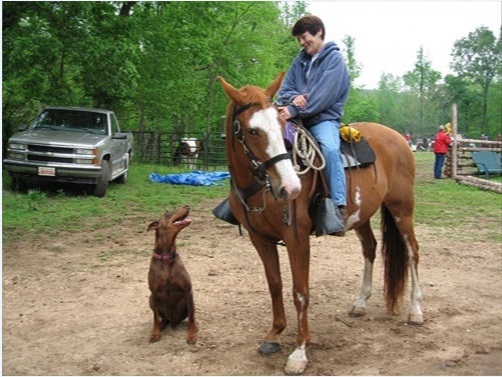}
    \end{minipage}

    \medskip
    \begin{minipage}[t]{0.325\linewidth}
    \includegraphics[width=\linewidth]{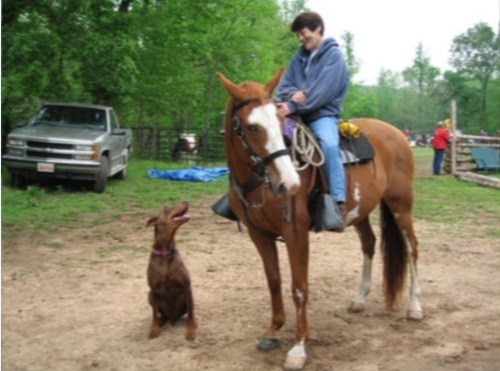}
    \end{minipage}\hfill
    \begin{minipage}[t]{0.325\linewidth}
    \includegraphics[width=\linewidth]{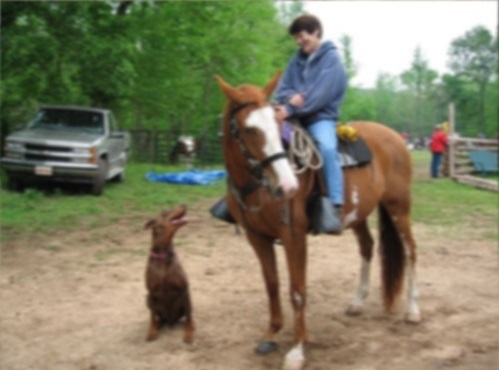}
    \end{minipage}\hfill
    \begin{minipage}[t]{0.325\linewidth}
    \includegraphics[width=\linewidth]{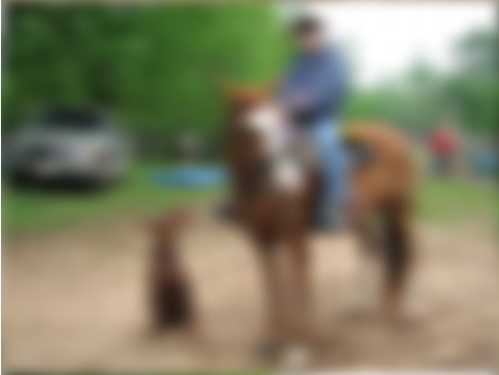}
    \end{minipage}
    \caption{\textbf{Top row from left to right.} Resolutions: full-spectrum, $R_{20}$ blurred with ($f_c = 20·\frac{w}{20})$, $R_{18}$: blurred with ($f_c = 18·\frac{w}{20})$. \textbf{Bottom row from left to right.} Resolutions: $R_{8}$ blurred with ($f_c = 8·\frac{w}{20})$, $R_{5}$: blurred with ($f_c = 5·\frac{w}{20})$, $R_{1}$: blurred with ($f_c = 1·\frac{w}{20})$}
    \label{fig: human and horse}
\end{figure*}

\textbf{Dataset.} The PASCAL Visual Object Classes 2007 (VOC2007) Dataset is considered as our image database that contains 9963 images that are categorized into 20 object classes as explained in the previous section. The data has been split into 5011 images for training and validation and 4952 test images. The distribution of images and objects by class is approximately equal across the training, validation and test datasets \cite{everingham2007pascal}.\\
Figure \ref{fig: human and horse} illustrates examples of an image from PASCAL VOC2007 database in the multiple levels of spatial frequencies ranging from low to high frequencies.

\subsection{Performance Evaluation of the Faster-RCNN on Different Resolutions}

To track the performance in the object detection, our performance evaluation metric is the mean Average Precision (mAP) which is a very common and popular performance measure in object detection tasks. The Average Precision is defined as the fraction of the images with a relevant detected object among all images with detected objects. In other words, the average precision is the area under the precision-recall curve for each categories of objects. The mean average precision is computed by taking the mean of the average precision for all category of objects \cite{everingham2010pascal, bayat2022humanahfe}.
\begin{figure}[t]
\begin{center}
\includegraphics[width=0.48\textwidth]{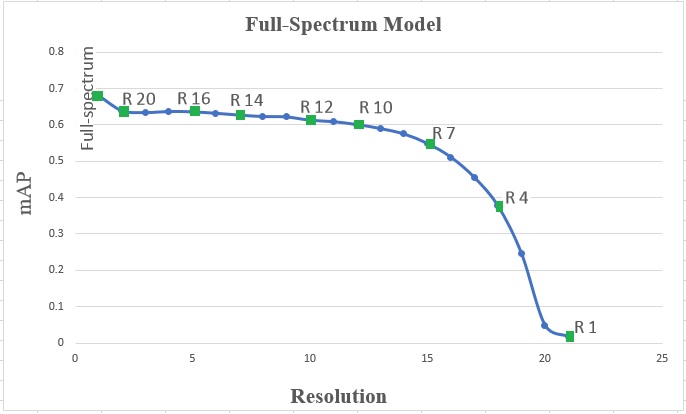}
\end{center}
\caption{Illustration of the performance of Faster- RCNN network on recognition of objects on the PASCAL VOC 2007 test dataset (4952 images) in multiple resolutions.}
\label{fig:Illustration of the performance multiple resolutions}
\end{figure}
Figure \ref{fig:Illustration of the performance multiple resolutions} shows the Faster-RCNN results of detection when tested on the PASCAL VOC2007 test data in various levels of resolution ($R_{1}, . . . , R_{20}$) as explained in the previous section. The results reveal that the performance of the Full-spectrum Model drops off quickly for low resolution images compared with the high resolution and full-spectrum images. This indicates that the representations that are learned for the object recognition in a deep neural network highly depend to the information from all spatial frequencies simultaneously. Hence, the lack of information from the specific scale negatively influences the recognition performance. However, the human visual system can detect objects in most of the resolution levels since the human brain provides representations of objects and scenes at multiple scales so that it can still interpret the scenes and objects even in a single level representation.\\
\begin{table*}[t]
\begin{center}
\caption{Performance comparison (mAP) of Faster- RCNN models trained on different resolutions for different resolution test cases created from PASCAL VOC2007 database}
\begin{tabular}{p{2.93cm}||p{2.25cm}|p{1.5cm}|p{1.5cm}|p{1.5cm}|p{1.5cm}}
\hline\hline
    Database-Resolution & \centering{Full-spectrum Model mAP(\%)} & \centering{$\frac{20}{20}$ -Model mAP(\%)} & \centering{$\frac{18}{20}$ -Model mAP(\%)} & \centering{$\frac{10}{20}$ -Model mAP(\%)} & $\frac{5}{20}$ -Model mAP(\%)\\ \hline
   \centering{Full-spectrum} & \centering{68.1\%} & \centering{\textbf{68.7\%}} & \centering{68.4\%} & \centering{67.6\%} & 61.6\% \\ \hline
    \centering{$R_{20}$} & \centering{63.9\%} & \centering{67.2\%} & \centering{\textbf{67.3\%}} & \centering{66.6\%} & 61.7\%   \\ \hline
    \centering{$R_{18}$} & \centering{63.7\%} & \centering{\textbf{67.5\%}} & \centering{67.3\%} & \centering{66.7\%} & 62.1\%  \\ \hline
    \centering{$R_{10}$} & \centering{60.1\%} & \centering{64.9\%} & \centering{65.3\%} & \centering{\textbf{65.9\%}} & 63.5\%  \\ \hline
    \centering{$R_5$} & \centering{45.5\%} & \centering{52.3\%} & \centering{51.2\%} & \centering{57.0\%} & \textbf{61.3\%}  \\ \hline \hline

\end{tabular}
\label{tb:Performance comparison}
\end{center}
\end{table*}

\subsection{Deploying a Multi-Resolution Faster-RCNN}
We propose a Multi-Resolution Faster-RCNN model that is made up of 5 end-to-end trained models on various resolution levels. A combination rule is applied during the test scheme such that a given image is passed through all 5 models and the best object recognition results are derived based on the combination rule that will be discussed shortly.

\begin{figure*}[t]
\begin{center}
\includegraphics[width=0.67\textwidth]{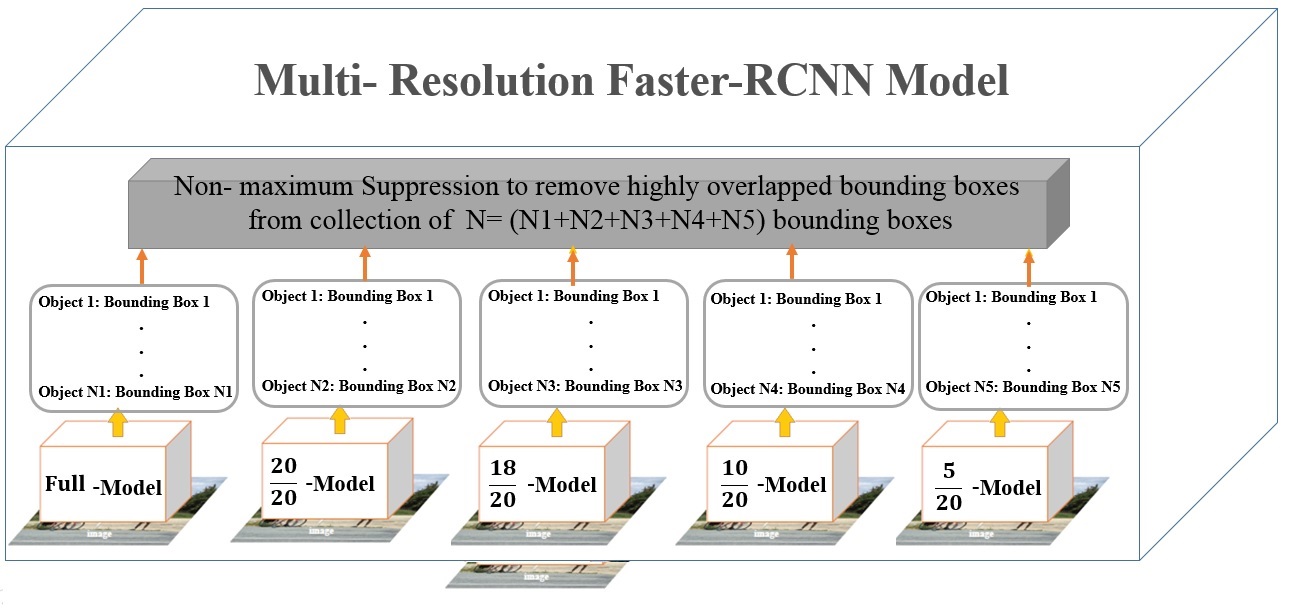}
\end{center}
\caption{The detection scheme in the Multi-Resolution Faster-RCNN.}
\label{fig:The detection scheme}
\end{figure*}

The combinational rule is adopted as an external module independent of the training scheme. To detect objects in a given input image to the Multi-Resolution Faster- RCNN model, all detections are collected from each of the five individual models. Each detected object is provided in the form of a bounding box and a score indicating the predicted probability of that bounding box belonging to an object class. The number of detections in each model may vary between 0 to 300, depending on the input image. Thus it is expected to have between 1 to 1500 proposed objects (presented as bounding box coordinates, predicted class score, and object class) for the combination of all five models. However, many of these bounding boxes highly overlap. For this, non-maximum suppression is used on the collection of all detected bounding boxes to reduce the redundancy. The Intersection-over-Union (IoU) threshold for non- maximum suppression is adopted at 0.7 to remove the redundant bounding boxes \cite{ren2016faster}. All remaining objects are proposed as the detection results of the Multi-Resolution Faster-RCNN model. Figure \ref{fig:The detection scheme} illustrates the detection scheme.

For evaluation of the the Multi-Resolution Faster-RCNN model, the process is implemented using the five models that were trained on PASCAL VOC2007 training/validation data (5K) in five different resolutions. Then, 20 test databases in 20 levels of resolutions were generated from the PASCAL VOC2007 test data (5k). For
each test database (corresponding to a certain resolution), the results of object detections from 5 models are obtained. Non-maximum suppression is adapted both based on the IoU threshold value of 0.7 as well as highest overlap with the ground truth bounding boxes. The detection results for five models and the Multi-resolution Faster-RCNN on 20 test databases (PASCAL VOC2007-$R_{1,...,20}$) are shown in Figure \ref{fig: The performance of multi-resolution}. The combined model, Multi-Resolution Faster-RCNN, outperforms all the models in detecting images in all ranges of resolutions. The results indicate the robustness, efficiency, and higher performance of the Multi-Resolution Faster-RCNN regardless of the resolution of the input image in comparison to the Faster-RCNN for object recognition.
\section{Conclusion and Future Work}
Inspired by the capability of human visual system for adapting to different resolution of images, in this work, we developed a multi-resolution deep object recognition framework which solves the issue of significant drop in the detection accuracy that happens for deep neural networks when trained and evaluated on different resolutions. This indicates that the
representations that are learned for the object recognition in deep neural network
highly depend to the information from all spatial frequencies simultaneously. Hence,
the lack of information from the certain scale negatively influence the recognition performance. To address this, we propose a Multi-Resolution Faster-RCNN model that is made up of 5 end to end
trained models on various resolution levels. The combination rule is applied during
the test scheme such that a given image is passed through all 5 models and the best
object recognition results are derived based on the combination rule. Our experiments show that 
the Multi-Resolution Faster-RCNN outperforms the original Faster-RCNN object detector in
detecting images in all ranges of resolutions. The results indicate the robustness,
efficiency, and higher performance of the Multi-Resolution Faster-RCNN regardless
of the resolution of the input image than Faster-RCNN for object recognition.


\bibliographystyle{IEEEtran}
\bibliography{IEEEabrv,references}

\EOD

\end{document}